\documentclass[runningheads]{llncs}

\usepackage{eccv}

\usepackage{eccvabbrv}

\usepackage{graphicx}
\usepackage{booktabs}

\usepackage[accsupp]{axessibility}  % Improves PDF readability for those with disabilities.
\usepackage{algorithm}
\usepackage{algpseudocode}

\usepackage[breaklinks,colorlinks,citecolor=eccvblue]{hyperref}
\usepackage{orcidlink}

\begin{document}

% ---------------------------------------------------------------
% TODO REVIEW: Replace with your title
\title{ClarifAI: Enhancing AI Interpretability and Transparency through Case-Based Reasoning and Ontology-Driven Approach for Improved Decision-Making} 

% TODO REVIEW: If the paper title is too long for the running head, you can set
% an abbreviated paper title here. If not, comment out.
\titlerunning{ClarifAI}

% TODO FINAL: Replace with your author list. 
% Include the authors' OCRID for the camera-ready version, if at all possible.
\author{Srikanth Vemula\inst{1}\inst{2}\orcidlink{0000-0001-6632-6749}}

% TODO FINAL: Replace with an abbreviated list of authors.
 \authorrunning{S.~Vemula}
% First names are abbreviated in the running head.
% If there are more than two authors, 'et al.' is used.

% TODO FINAL: Replace with your institution list.
\institute{Human-Centric eXplainable Intelligence Lab (hcXiL) \and Computer Science Department(CSCI), Saint John's University (CSBSJU)
\email{svemula001@csbsju.edu}}

\maketitle

\begin{abstract}
This study introduces Clarity and Reasoning Interface for Artificial Intelligence (ClarifAI), a novel approach designed to augment the transparency and interpretability of artificial intelligence (AI) in the realm of improved decision making. Leveraging the Case-Based Reasoning (CBR) methodology and integrating an ontology-driven approach, ClarifAI aims to meet the intricate explanatory demands of various stakeholders involved in AI-powered applications. The paper elaborates on ClarifAI’s theoretical foundations, combining CBR and ontologies to furnish exhaustive explanation mechanisms. It further elaborates on the design principles and architectural blueprint, highlighting ClarifAI's potential to enhance AI interpretability across different sectors and its applicability in high-stake environments. This research delineates the significant role of ClarifAI in advancing the interpretability of AI systems, paving the way for its deployment in critical decision-making processes.
  \keywords{ClarifAI\and Case-Based Reasoning\and Ontology\and AI Interpretability\and Decision-Making}
\end{abstract}

\section{Introduction}
\label{sec:intro}

In the era of rapid technological advancement, artificial intelligence (AI) has emerged as a cornerstone of innovation across various sectors, revolutionizing the way decisions are made. However, the increasing complexity of AI systems and their opaque decision-making processes have raised significant concerns regarding transparency and interpretability. These concerns are especially pronounced in scenarios where decisions have far-reaching implications, necessitating a bridge between AI's complex mechanisms and human understanding.

To address these challenges, this paper introduces ClarifAI, an novel platform designed to enhance the interpretability and transparency of AI-assisted decision-making. At the heart of ClarifAI lies the integration of Case-Based Reasoning (CBR) and an ontology-driven framework. This integration is pivotal, as it not only leverages the explanatory power of CBR but also harnesses the structured knowledge representation offered by ontologies. The fusion of these methodologies enables ClarifAI to provide nuanced and contextually relevant explanations tailored to the needs of diverse stakeholders involved in AI applications.

The objective of this paper is twofold. First, it aims to detail the theoretical underpinnings of ClarifAI, illustrating how the synergistic use of CBR and ontology-based approaches can lead to a more interpretable and transparent AI. Second, the paper explores the design principles and architectural framework of ClarifAI, highlighting its potential to set new standards for AI interpretability across different sectors. Through its focus on enhancing decision-making in high-stakes environments, ClarifAI represents a significant step forward in addressing the critical need for accountable and understandable AI systems.

This introduction sets the stage for a comprehensive exploration of ClarifAI, beginning with an overview of the current landscape of AI interpretability and transparency, followed by a detailed exposition of ClarifAI's conceptual framework, design principles, and potential impact on AI-assisted decision-making processes and finally concludes with conclusion and future work.

\section{Related Work}

\subsection{Overview of the Current Landscape of AI Interpretability and Transparency}
The quest for interpretability and transparency in artificial intelligence (AI) systems has become a paramount concern in the field, given the widespread adoption of AI in critical decision-making processes. Interpretability refers to the extent to which a human can understand the cause of a decision made by an AI system, while transparency involves the accessibility of an AI system's operations to human inspection. The significance of these aspects cannot be overstated, especially as AI systems increasingly influence sectors such as healthcare, finance, and legal systems, where decisions impact human lives directly.

Recent literature underscores the growing emphasis on developing methodologies and frameworks to enhance AI interpretability and transparency. Doshi-Velez and Kim provided a comprehensive overview of interpretability, identifying it as a key factor in building trust and reliability in AI systems~\cite{DoshiVelez2017TowardsAR}. Similarly, Lipton highlighted the importance of transparency, noting that it facilitates better understanding, debugging, and trust in AI models~\cite{Lipton16}.

The emergence of explainable AI (XAI) frameworks aims to address these concerns by making AI decision-making processes more understandable to humans. Guidotti et al.reviewed various approaches to XAI, categorizing them into model-agnostic and model-specific techniques, and emphasizing the importance of explanations in AI systems' acceptance and ethical use~\cite{Guidotti2018ASO}. Moreover, the integration of Case-Based Reasoning (CBR) and ontology-driven approaches has been identified as a promising direction for enhancing AI interpretability. For instance, Sørmo et al. discussed how CBR can contribute to explainability by leveraging past cases to justify decisions, thereby providing a more intuitive understanding of AI processes~\cite{SormoEtAl2005}.

Despite these advancements, challenges remain in achieving optimal transparency and interpretability. Gunning and Aha outlined ongoing efforts and future directions in XAI, suggesting that balancing the complexity of explanations with human cognitive capabilities is crucial~\cite{Gunning_Aha_2019}. Furthermore, the development of standards and benchmarks for evaluating AI interpretability and transparency is an active area of research, as indicated by Sokol and Flach, who argued for the necessity of robust metrics to assess the effectiveness of XAI solutions~\cite{SokolFlach20}.

In summary, the current landscape of AI interpretability and transparency is characterized by a heightened awareness of the need for explainable and transparent AI systems. Advances in XAI, including the use of CBR and ontologies, offer promising avenues for addressing these needs. However, the field continues to face challenges in developing universally accepted standards and methodologies that can adequately bridge the gap between AI complexity and human understanding.

\subsection{Case-Based Reasoning (CBR) in Interpretability}
Case-Based Reasoning (CBR) stands out as a significant paradigm in enhancing interpretability and explainability in artificial intelligence (AI). CBR operates on the principle of solving new problems by referring to similar past cases and learning from their outcomes.This approach inherently supports interpretability, as the decision-making process is grounded in concrete examples that are understandable to humans.

Kolodner laid the foundational framework for CBR, illustrating its potential for problem-solving by analogy and experience reuse, which are intuitively accessible to human reasoning~\cite{Kolodner93}. More recently, Leake and Sooriamurthi emphasized the natural alignment of CBR with human explanatory patterns, noting that CBR's reliance on historical data makes its decisions inherently more explainable compared to other AI methods ~\cite{LeakeSooriamurthi20}. This is because CBR provides a narrative through its case retrieval and adaptation steps, offering a story-like explanation for its decisions.

\subsection{Ontology-Driven Approaches in Explainability}
Ontologies, as formal representations of knowledge within a domain, play a crucial role in enhancing the explainability of AI systems. By structuring domain knowledge into hierarchies and relationships, ontologies facilitate a deeper understanding of the decision-making context, thereby improving the explainability of AI systems.

Gómez-Pérez, Fernández-López, and Corcho highlighted the importance of ontologies in knowledge management and the semantic web, underscoring their utility in making AI systems more interpretable and aligned with human conceptualizations~\cite{GmezPrez2004OntologicalEW}. In the context of explainable AI, Ibrahim et al. discussed how ontology-driven approaches could structure domain knowledge to make AI decisions more transparent and justifiable~\cite{Ibrahim13}. Ontologies enable the explicit representation of the rationale behind AI decisions, linking them to well-defined concepts and relationships that are comprehensible to users.

\subsection{Synergy of CBR and Ontology-Driven Approaches}
The integration of CBR and ontology-driven approaches offers a promising avenue for advancing interpretability and explainability in AI. This synergy leverages the narrative power of CBR and the structured knowledge representation of ontologies to produce AI systems whose decisions are both grounded in real-world examples and deeply rooted in domain-specific knowledge.

Haque et al. explored this idea of integration, which enhances the semantic richness of the cases and facilitates better understanding and reasoning about decisions~\cite{HaqueEtAl2022}. Furthermore, Ahmed et al.demonstrated the application of this combined approach in technical diagnosis systems, where the integration significantly improved the explainability of the system's recommendations, making it easier for technicians to understand and trust the AI's advice~\cite{AhmedEtAl2010}.

In conclusion, CBR and ontology-driven approaches individually contribute significantly to the interpretability and explainability of AI systems. When combined, they offer a robust framework for developing AI solutions that are both transparent and easily understandable by humans. This literature review underscores the importance of these approaches in bridging the gap between complex AI algorithms and human-centric explanations, paving the way for more accountable and trustworthy AI systems.

\section{ClarifAI: Clarity and Reasoning Interface for Artificial Intelligence}
The conceptual framework of ClarifAI represents an innovative strategy designed to significantly enhance the interpretability and transparency of decisions made by artificial intelligence (AI) systems, addressing an essential demand for comprehensible AI methodologies across varied application domains. Central to ClarifAI’s design is the synergistic use of Case-Based Reasoning (CBR) and ontology-based approaches, which together form the core components and operational flow of the platform. This integration not only leverages the explanatory power inherent in CBR—where decisions are illuminated through similarities with past cases—but also enriches explanations with the structured knowledge encapsulated within ontologies. Such a combination ensures that explanations are not only relatable but also deeply informative, rooted in the specific semantics of the domain at hand. The theoretical underpinnings of ClarifAI highlight the importance of this synergy, offering a solid foundation upon which the platform’s design principles are established. The architectural framework of ClarifAI, meticulously crafted around these principles, ensures a robust, scalable, and user-friendly system. This section aims to unravel the intricacies of ClarifAI's conceptual framework, detailing its core components, operational workflow, and the theoretical rationale behind its innovative approach to AI interpretability and transparency. Through a deep dive into the design principles and architectural framework, we illuminate the pathways through which ClarifAI seeks to revolutionize the landscape of explainable AI.
\subsection{ClarifAI's Conceptual Framework}
ClarifAI's conceptual framework is designed to address the pressing need for interpretability and transparency in AI-assisted decision-making processes. The framework integrates Case-Based Reasoning (CBR) with an ontology-driven approach, leveraging the strengths of both to create a system that can provide comprehensible and contextually relevant explanations for its decisions. This section provides a detailed exposition of the conceptual underpinnings of ClarifAI, illustrating how it operates and the rationale behind its design.

\subsubsection{Core Components:}
Case-Based Reasoning (CBR): At the core of ClarifAI is the CBR mechanism, which solves new problems by referring to similar past cases. This approach is inherently interpretable, as it allows for decisions to be explained through examples that are familiar to the user.

Ontology-Driven Framework: ClarifAI incorporates a rich ontology-driven framework that organizes domain knowledge in a structured manner. Ontologies facilitate understanding by providing a clear representation of the relationships and hierarchies between different concepts within the domain.

Explanatory Interface: A key feature of ClarifAI is its explanatory interface, which presents the rationale behind decisions in an accessible and user-friendly manner. This interface uses the information from the CBR system and the ontology framework to generate explanations that are both informative and easy to understand.

\subsubsection{Operational Flow:}

Input Processing: When a new problem is presented to ClarifAI, the system first processes the input to understand the context and specifics of the request.

Case Retrieval: The CBR system then searches the case database to find past cases that are similar to the current problem. This involves comparing the new problem with stored cases based on their features and the outcomes of those cases.

Case Adaptation: Once a relevant case (or cases) is identified, ClarifAI adapts the solution of the past case to fit the specifics of the new problem. This step may involve modifying the solution based on the differences between the cases.

Ontology Mapping: Parallel to the CBR process, the ontology framework is consulted to enrich the decision-making process with domain-specific knowledge. This step ensures that the solution is not only based on past cases but is also grounded in the conceptual structure of the domain.

Explanation Generation: With the solution derived from the CBR process and contextualized with the ontology framework, the explanatory interface generates a comprehensive explanation. This explanation details why the solution was chosen and how it relates to past cases and domain knowledge.

Output Presentation: Finally, the solution and its explanation are presented to the user. The explanation aims to make the decision-making process transparent and understandable, fostering trust in the system.

The following diagram and flowchart illustrate the operational flow of ClarifAI, from input processing to the presentation of the solution and its explanation:

\subsubsection{Theoretical Underpinnings of ClarifAI: Synergistic Use of CBR and Ontology-Based Approaches}
\subsubsection{Case-Based Reasoning (CBR) Integration}

The integration of Case-Based Reasoning (CBR) into ClarifAI provides a foundation for interpretability by leveraging the power of analogical problem-solving. CBR operates on the principle that similar problems have similar solutions, making the decision-making process inherently more understandable to users. 
\begin{algorithm}
\caption{Retrieve Similar Case Algorithm}\label{alg:retrieve_similar_case}
\begin{algorithmic}
\Require $new\_case, case\_database$
\Ensure $most\_similar\_case$
\State $most\_similar\_case \gets None$
\State $highest\_similarity \gets -1$
\ForAll{$(case, solution) \in case\_database$}
    \State $similarity \gets compute\_similarity(new\_case, case)$
    \If{$similarity > highest\_similarity$}
        \State $highest\_similarity \gets similarity$
        \State $most\_similar\_case \gets (case, solution)$
    \EndIf
\EndFor
\State \Return $most\_similar\_case$
\end{algorithmic}
\end{algorithm}
This is because CBR relies on historical data and past cases to reason about new situations, offering a narrative or story that explains how a particular decision was reached. For instance, when faced with a new problem, ClarifAI retrieves and adapts solutions from similar past cases, providing users with concrete examples to justify its decisions. This approach not only makes the AI's thought process more transparent but also aligns closely with human cognitive processes, as humans often rely on past experiences to make decisions.To demonstrate this a Retrieve Similar Case algorithm pseudo code is presented in algorithm 1.

\subsubsection{Ontology-Based Framework Enhancement}
On the other hand, the ontology-driven framework of ClarifAI organizes and structures domain knowledge, enabling the system to reason with a deep understanding of the context in which decisions are made.
\begin{algorithm}
\caption{Generate Explanation Algorithm}\label{alg:generate_explanation}
\begin{algorithmic}
\Require $case\_solution, domain\_ontology$
\Ensure $explanation$
\State $explanation\_components \gets$ empty list
\ForAll{$concept \in case\_solution['concepts\_involved']$}
    \State $concept\_definition \gets domain\_ontology.get\_concept\_definition(concept)$
    \State append $concept: concept\_definition$ to $explanation\_components$
\EndFor
\State $explanation \gets$ concatenate $explanation\_components$ with spaces
\State \Return $explanation$
\end{algorithmic}
\end{algorithm}
Ontologies define a set of concepts and categories in a subject area or domain that shows their properties and the relations between them. By integrating ontologies, ClarifAI can interpret and classify the input data more accurately, ensuring that its decisions are grounded in a comprehensive understanding of the domain. This structured knowledge representation facilitates the generation of explanations that are not only relevant but also contextually rich, making it easier for users to understand the rationale behind AI decisions.To illustrate how CBR is integrated with ontology-driven framework which helps to generate explanation is shown in the Generate Explanation Algorithm in algorithm 2.

\subsubsection{Synergistic Benefits}
The synergy between CBR and ontology-driven framework approaches enhances ClarifAI's ability to provide interpretable and transparent AI solutions. While CBR brings interpretability through analogical reasoning with past cases, ontologies contribute to transparency by structuring domain knowledge. 
\begin{algorithm}
\caption{Clarify Decision Using CBR and Ontology}\label{alg:clarify_decision}
\begin{algorithmic}
\Require $new\_case, case\_database, domain\_ontology$
\Ensure $decision\_details$
\Function{clarify\_decision}{$new\_case, case\_database, domain\_ontology$}
    \State $(similar\_case, solution) \gets \Call{retrieve\_similar\_case}{new\_case, case\_database}$
    \State $explanation \gets \Call{generate\_explanation}{solution, domain\_ontology}$
    \State $decision\_details \gets \{\text{'similar\_case'}: similar\_case, \text{'solution'}: solution, \text{'explanation'}: explanation\}$
    \State \Return $decision\_details$
\EndFunction
\end{algorithmic}
\end{algorithm}
This combination allows ClarifAI to deliver explanations that are both grounded in practical examples (through CBR) and deeply rooted in the specificities of the domain (through ontologies). As a result, the platform can cater to the complex explanation needs of diverse stakeholders, making AI-assisted decisions more accessible and trustworthy.The above algorithm 3 shows how the syngery of CBR and ontology-driven framework helps ClarifAI to make decisions is shown.
\subsubsection{Design Principles and Architectural Framework of ClarifAI}
The architectural and design principles of ClarifAI is underpinned by emphasizing on interpretability and transparency, tailored to enhance user engagement and satisfaction across varied application domains:

User-Centric Design: ClarifAI prioritizes the user experience, crafting explanations that are accessible and meaningful to individuals regardless of their technical background. This approach ensures that the platform's insights are universally comprehensible, fostering a broader understanding and trust in AI decisions.

Flexible Adaptation: Acknowledging the unique demands of different fields, ClarifAI showcases exceptional adaptability. It allows for the customization of its Case-Based Reasoning (CBR) and ontology frameworks to align with the nuanced requirements of specific sectors, enhancing the relevance and applicability of its explanations.

Scalable Architecture: Anticipating the exponential growth in data and the complexity of decisions, the architecture of ClarifAI is designed for scalability. It efficiently supports the expansion of its case and ontology databases, ensuring sustained high performance even as system demands escalate.

Inherent Transparency: At its core, ClarifAI embodies transparency, with a built-in capacity to make each step of the decision-making process clear and traceable. This fundamental transparency assures that the system’s operations are open and understandable, instilling confidence in its outputs.

\subsubsection{Architectural Framework}
The architecture of ClarifAI seamlessly integrates Case-Based Reasoning (CBR) with an ontology-driven framework to deliver nuanced and transparent AI-assisted decision-making. This integration is visualized in a high-level architecture diagram that illustrates the flow from data input through to the generation of explainable outputs:
\begin{figure}[tb]
  \centering
  \includegraphics[height=6.5cm]{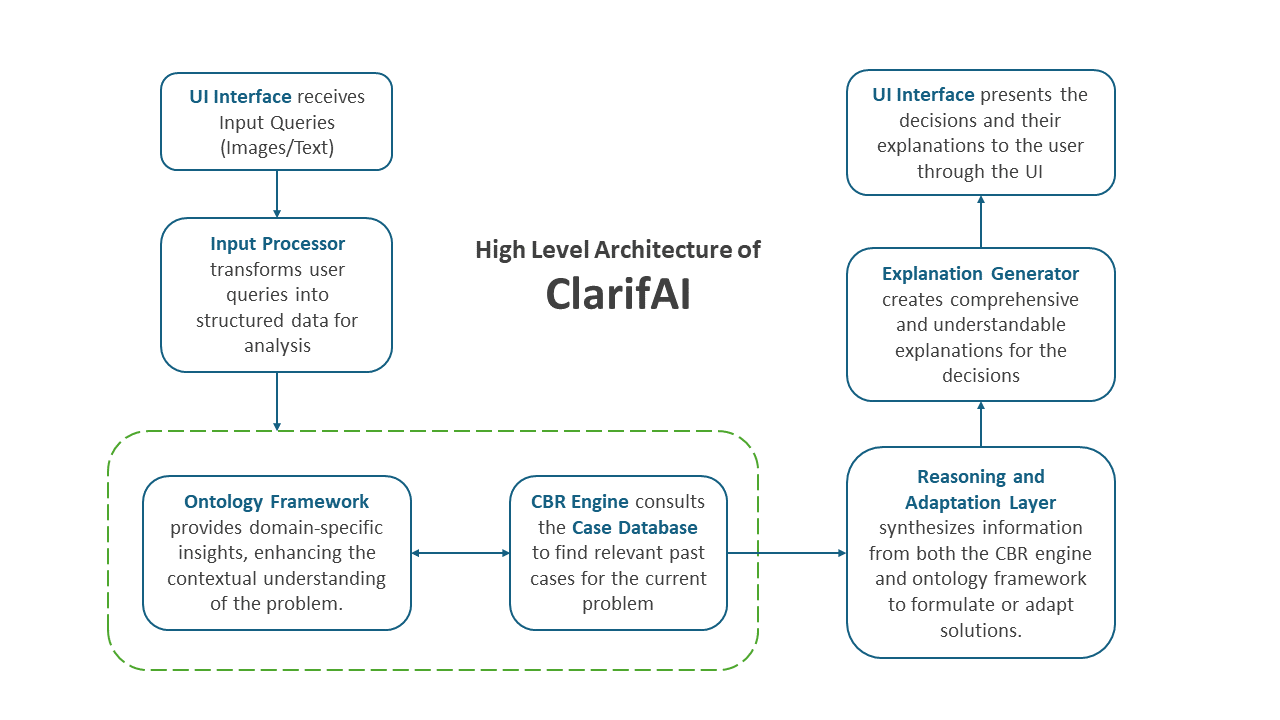}
  \caption{High level architecture diagram for ClarifAI, integrating Case-Based Reasoning (CBR) and ontology-driven approaches for AI-assisted decision-making.This architecture encapsulates the synergistic operation of CBR and ontology-driven approaches within ClarifAI, aimed at enhancing AI interpretability and transparency across decision-making processes.
  }
  \label{fig:example}
\end{figure}
Data Input Layer: This initial layer captures and preprocesses data from diverse sources, preparing it for analysis.
CBR Engine: Central to ClarifAI, the CBR engine retrieves relevant past cases from a comprehensive database, utilizing similarities to inform current decision-making.
Ontology Framework: Running parallel to the CBR engine, the ontology framework structures domain knowledge, enriching the decision-making process with contextual depth and clarity.
Explanation Generator: This component synthesizes insights from both the CBR engine and the ontology framework, crafting tailored explanations that elucidate the AI’s reasoning in user-friendly terms.
Output Interface: The final layer presents the generated explanations to users, ensuring they are accessible and actionable.

Together, these layers form a robust framework that supports ClarifAI's objectives of enhancing AI interpretability and transparency. By detailing the theoretical underpinnings and exploring the design principles and architectural framework, this paper highlights ClarifAI's potential to set new standards for AI interpretability across various sectors, offering a path toward more accountable and understandable AI systems.

\section{Potential Impact of ClarifAI on AI-Assisted Decision-Making Processes
}
\label{sec:blind}
The introduction of ClarifAI, with its unique integration of Case-Based Reasoning (CBR) and ontology-driven frameworks, promises to significantly impact the landscape of AI-assisted decision-making processes. By prioritizing interpretability and transparency, ClarifAI addresses a critical gap in current AI applications, fostering trust and understanding among users. This section explores the potential impact of ClarifAI across various dimensions of AI-assisted decision-making.

\subsubsection{Enhancing Trust in AI Systems}
One of the most profound impacts of ClarifAI is its potential to enhance trust in AI systems. Trust is foundational to the adoption and effective use of AI technologies, especially in sectors where decisions have significant implications, such as healthcare, finance, and public policy~\cite{Ribeiro16}. By providing clear, understandable explanations for its decisions, ClarifAI enables users to comprehend the rationale behind AI-generated outcomes, thereby increasing their confidence in the system. This trust is essential for users to rely on AI assistance for critical decision-making tasks, knowing that the system's recommendations are both reliable and transparent.

\subsubsection{Improving Decision Quality and Efficiency}
ClarifAI's approach to integrating past case solutions and domain-specific knowledge can significantly improve the quality and efficiency of decision-making processes. In scenarios where decision-makers are faced with complex, multifaceted problems, ClarifAI can quickly provide relevant, contextually informed solutions by drawing on a vast repository of past cases and structured domain knowledge. This not only speeds up the decision-making process but also ensures that decisions are informed by a comprehensive understanding of the domain, potentially leading to higher-quality outcomes.

\subsubsection{Facilitating Compliance and Accountability}
In many regulated industries, decisions made by AI systems must comply with existing legal and ethical standards. ClarifAI's transparent decision-making process, supported by detailed explanations, can greatly facilitate compliance with such standards. By documenting how and why decisions were made, ClarifAI provides a clear audit trail that can be reviewed by regulatory bodies, ensuring that AI-assisted decisions adhere to required norms and regulations. This transparency is crucial for accountability, allowing organizations to demonstrate the fairness and objectivity of their AI-driven processes~\cite{Binns2017FairnessIM}.

\subsubsection{Democratizing Access to AI Technologies}
ClarifAI has the potential to democratize access to advanced AI technologies by making them more understandable and accessible to a broader audience. Often, the complexity of AI systems acts as a barrier to entry for individuals and organizations with limited technical expertise. However, by providing intuitive, case-based explanations and leveraging structured domain knowledge, ClarifAI makes it easier for non-experts to engage with and benefit from AI technologies. This democratization can lead to more widespread adoption and innovative applications of AI across different sectors.

\subsubsection{Fostering Innovation in AI Applications}
Finally, ClarifAI's novel approach to interpretability and transparency can spur innovation in AI applications. By setting new standards for explainability, ClarifAI challenges the field to rethink how AI systems are designed and implemented. This can inspire the development of new AI models and applications that prioritize user understanding and ethical considerations, leading to more responsible and impactful uses of AI technology~\cite{Hoffman2018MetricsFE}.

In summary, ClarifAI's impact on AI-assisted decision-making processes is multifaceted, with the potential to enhance trust, improve decision quality, ensure compliance, democratize access, and foster innovation. By addressing the critical need for interpretable and transparent AI, ClarifAI sets the stage for a future where AI technologies are not only powerful but also aligned with human values and understanding.

\section{Conclusion}
The development and deployment of ClarifAI mark a significant advancement in the quest for more interpretable and transparent AI-assisted decision-making processes. By seamlessly integrating Case-Based Reasoning (CBR) and ontology-driven frameworks, ClarifAI addresses the crucial need for AI systems that users can trust and understand. This paper has detailed the theoretical underpinnings of ClarifAI, highlighting how the synergistic use of CBR and ontologies can lead to enhanced interpretability and transparency in AI. Furthermore, it has explored the design principles and architectural framework that underlie ClarifAI, showcasing its potential to set new standards for AI interpretability across different sectors.

ClarifAI's impact is poised to be profound, offering the promise of improved trust in AI systems, enhanced decision quality and efficiency, facilitated compliance and accountability, democratized access to AI technologies, and the fostering of innovation in AI applications. By making AI decisions more understandable and accountable, ClarifAI not only enhances the user experience but also paves the way for more ethical and responsible AI applications.
\subsubsection{Future Work:}
In the context of human-robot interaction (HRI), ClarifAI's potential to improve decision-making processes through enhanced interpretability and transparency is significant. As robots become more autonomous, their decisions—ranging from navigation to interaction strategies—must be interpretable to their human counterparts. Future work will explore how ClarifAI can be implemented in HRI systems to generate explanations for robot decisions that are easily understandable by humans. This includes tailoring the CBR component to leverage cases specific to HRI scenarios and enriching the ontology with concepts relevant to robotics and human factors.

Experimental studies involving humans interacting with robots, where ClarifAI provides explanations for the robot's decisions, will be crucial. These studies will evaluate the impact of explanations on human trust, satisfaction, and perceived effectiveness of the robot, informing further refinements to the ClarifAI framework for HRI applications.

In conclusion, ClarifAI represents a significant step forward in making AI more interpretable and transparent, fostering a greater understanding and trust in AI technologies. The potential for future enhancements and applications of ClarifAI is vast, with the promise of transforming AI-assisted decision-making processes across a wide range of domains. As the platform evolves, it will continue to contribute to the development of AI systems that are not only powerful and efficient but also ethical, understandable, and aligned with human values.

% ---- Bibliography ----
%
% BibTeX users should specify bibliography style 'splncs04'.
% References will then be sorted and formatted in the correct style.
%
\bibliographystyle{splncs04}
\bibliography{main}

\begin{thebibliography}{10}
\providecommand{\url}[1]{\texttt{#1}}
\providecommand{\urlprefix}{URL }
\providecommand{\doi}[1]{https://doi.org/#1}

\bibitem{AhmedEtAl2010}
Ahmed, M., Begum, S., Olsson, E., Xiong, N., Funk, P.: Case-based reasoning for medical and industrial decision support systems. In: Successful Case-based Reasoning Applications - I, Studies in Computational Intelligence. vol.~305, pp. 7--52. Springer, Berlin, Heidelberg (2010). \doi{10.1007/978-3-642-14078-5_2}

\bibitem{Binns2017FairnessIM}
Binns, R.: Fairness in machine learning: Lessons from political philosophy. Decision-Making in Computational Design \& Technology eJournal  (2017), \url{https://api.semanticscholar.org/CorpusID:3315224}

\bibitem{DoshiVelez2017TowardsAR}
Doshi-Velez, F., Kim, B.: Towards a rigorous science of interpretable machine learning. arXiv: Machine Learning  (2017), \url{https://api.semanticscholar.org/CorpusID:11319376}

\bibitem{GmezPrez2004OntologicalEW}
G{\'o}mez-P{\'e}rez, A., Fern{\'a}ndez-L{\'o}pez, M., Corcho, {\'O}.: Ontological engineering: With examples from the areas of knowledge management, e-commerce and the semantic web. In: Advanced Information and Knowledge Processing (2004), \url{https://api.semanticscholar.org/CorpusID:26784846}

\bibitem{Guidotti2018ASO}
Guidotti, R., Monreale, A., Turini, F., Pedreschi, D., Giannotti, F.: A survey of methods for explaining black box models. ACM Computing Surveys (CSUR)  \textbf{51},  1 -- 42 (2018), \url{https://api.semanticscholar.org/CorpusID:3342225}

\bibitem{Gunning_Aha_2019}
Gunning, D., Aha, D.: Darpa’s explainable artificial intelligence (xai) program. AI Magazine  \textbf{40}(2),  44--58 (Jun 2019). \doi{10.1609/aimag.v40i2.2850}

\bibitem{HaqueEtAl2022}
Haque, A.K.M.B., Arifuzzaman, B.M., Siddik, S.A.N., Kalam, A., Shahjahan, T.S., Saleena, T.S., Alam, M., Islam, M.R., Ahmmed, F., Hossain, M.J.: Semantic web in healthcare: A systematic literature review of application, research gap, and future research avenues. International Journal of Clinical Practice  \textbf{2022},  1--27 (2022). \doi{10.1155/2022/6807484}

\bibitem{Hoffman2018MetricsFE}
Hoffman, R.R., Mueller, S.T., Klein, G., Litman, J.: Metrics for explainable ai: Challenges and prospects. ArXiv  \textbf{abs/1812.04608} (2018), \url{https://api.semanticscholar.org/CorpusID:54577009}

\bibitem{Ibrahim13}
Ibrahim, A., Hashi, H., Ali, A.: Ontology-driven information retrieval for healthcare information system : A case study. International Journal of Network Security \& Its Applications  \textbf{5},  61--69 (2013). \doi{10.5121/ijnsa.2013.5105}

\bibitem{Kolodner93}
Kolodner, J.: Case-Based Reasoning. Morgan Kaufmann (1993)

\bibitem{LeakeSooriamurthi20}
Leake, D.B., Sooriamurthi, R.: Case-Based Reasoning: Experiences, Lessons, and Future Directions. Menlo Park: AAAI Press/MIT Press (1996), \url{https://homes.luddy.indiana.edu/leake/papers/p-96-01.pdf}

\bibitem{Lipton16}
Lipton, Z.C.: The mythos of model interpretability. Communications of the ACM  \textbf{61},  36 -- 43 (2016), \url{https://api.semanticscholar.org/CorpusID:5981909}

\bibitem{Ribeiro16}
Ribeiro, M.T., Singh, S., Guestrin, C.: "why should i trust you?": Explaining the predictions of any classifier. In: Proceedings of the 22nd ACM SIGKDD International Conference on Knowledge Discovery and Data Mining. p. 1135–1144. KDD '16, Association for Computing Machinery, New York, NY, USA (2016). \doi{10.1145/2939672.2939778}

\bibitem{SokolFlach20}
Sokol, K., Flach, P.A.: Explainability fact sheets: {A} framework for systematic assessment of explainable approaches. CoRR  \textbf{abs/1912.05100} (2019), \url{http://arxiv.org/abs/1912.05100}

\bibitem{SormoEtAl2005}
Sørmo, F., Cassens, J., Aamodt, A.: Explanation in case-based reasoning–perspectives and goals. Artificial Intelligence Review  \textbf{24},  109--143 (2005). \doi{10.1007/s10462-005-4607-7}

\end{thebibliography}
\end{document}